\def\year{2021}\relax
\documentclass[letterpaper]{article} % DO NOT CHANGE THIS
\usepackage{aaai21}  % DO NOT CHANGE THIS
\usepackage{times}  % DO NOT CHANGE THIS
\usepackage{helvet} % DO NOT CHANGE THIS
\usepackage{courier}  % DO NOT CHANGE THIS
\usepackage[hyphens]{url}  % DO NOT CHANGE THIS
\usepackage{graphicx} % DO NOT CHANGE THIS
\usepackage{booktabs}
\usepackage{subcaption}
\usepackage{xcolor}
\usepackage{tabularx}
\usepackage{fdsymbol}

\usepackage{natbib}  % DO NOT CHANGE THIS AND DO NOT ADD ANY OPTIONS TO IT
\usepackage{caption} % DO NOT CHANGE THIS AND DO NOT ADD ANY OPTIONS TO IT
\newcommand{\modelname}{\textsc{GAP~}}
\newcommand{\modelnamelm}{\textsc{GAP~Model}}
\newcommand{\colpred}{CPred}
\newcommand{\fullcolpred}{Column Prediction~}
\newcommand{\colrec}{CRec}
\newcommand{\fullcolrec}{Column Recovery~}
\newcommand{\gensql}{GenSQL}
\newcommand{\fullgensql}{SQL Generation~}

\newcolumntype{L}[1]{>{\raggedright\let\newline\\\arraybackslash\hspace{0pt}}m{#1}}

\title{Learning Contextual Representations for Semantic Parsing with Generation-Augmented Pre-Training}

\author {
    Peng Shi\textsuperscript{\rm 1}\footnote{Work done while at AWS AI Labs.},
    Patrick Ng\textsuperscript{\rm 2},
    Zhiguo Wang\textsuperscript{\rm 2},
    Henghui Zhu\textsuperscript{\rm 2},
    Alexander Hanbo Li\textsuperscript{\rm 2}, \\
    Jun Wang\textsuperscript{\rm 2},
    Cicero Nogueira dos Santos\textsuperscript{\rm 2},
    Bing Xiang\textsuperscript{\rm 2}  \\
}
\affiliations {
    \textsuperscript{\rm 1} University of Waterloo,
    \textsuperscript{\rm 2} AWS AI Labs \\
    peng.shi@uwaterloo.ca,
    \{patricng,zhiguow,henghui,hanboli,juwanga,cicnog,bxiang\}@amazon.com \\
}
\begin{document}

\maketitle

\begin{abstract}

Most recently, there has been significant interest in learning contextual representations for various NLP tasks,
by leveraging large scale text corpora to train large neural language models with self-supervised learning objectives, such as Masked Language Model~(MLM).
However, based on a pilot study, we observe three issues of existing general-purpose language models when they are applied to text-to-SQL semantic parsers:
fail to detect column mentions in the utterances, fail to infer column mentions from cell values, and fail to compose complex SQL queries.
To mitigate these issues,
we present a model pre-training framework, \textbf{G}eneration-\textbf{A}ugmented \textbf{P}re-training~(\textsc{GAP}), that jointly learns representations of natural language utterances and table schemas by leveraging generation models to generate pre-train data.
\modelnamelm\footnote{This refers to the language models that are pre-trained with GAP framework.} is trained on 2M utterance-schema pairs and 30K utterance-schema-SQL triples, whose utterances are produced by generative models.
Based on experimental results, neural semantic parsers that leverage \modelnamelm~as a representation encoder obtain new state-of-the-art results on both \textsc{Spider} and \textsc{Criteria-to-SQL} benchmarks.

\end{abstract}

\section{Introduction}

Recently, deep contextual language models~\cite{devlin2018bert, liu2019roberta,lewis2019bart, dong2019unified, raffel2019exploring} have shown their effective modeling ability for text, achieving state-of-the-art results in series of NLP tasks.
These models capture the syntactic and semantic information of the input text, generating fine-grained contextual embeddings, which can be easily applied to downstream models.
Despite the success of large scale pre-trained language models on various tasks, 
it is less clear how to extend them to semantic parsing tasks such as text-to-SQL~\cite{warren1982efficient, popescu2003towards, popescu2004modern, li2006constructing},
which requires joint reasoning of the natural language utterance and structured database schema information.
Recent work~\cite{guo2019towards,wang2019rat, bogin2019representing, bogin2019global} shows that with more powerful pre-trained language models, the highly domain-specific semantic parsers can be further improved, even though these language models are trained for pure text encoding.

\begin{table}[t]
  \centering \small
    \begin{tabularx}{\linewidth}{L{0.97 \linewidth}}
    \toprule
    \textbf{Pain Point 1}: Fail to match and detect the column mentions. \\
    \textbf{Utterance}: Which professionals live in a city containing the substring 'West'? List his or her role, street, \textit{city} and state. \\
    \textbf{Prediction}: \texttt{SELECT role\_code, street, state FROM Professionals WHERE city LIKE '\%West\%'} \\
    \textbf{Error}: Missing column \texttt{city} in \texttt{SELECT} clause. \\

    \midrule
    \textbf{Pain Point 2}: Fail to infer columns based on cell values. \\
    \textbf{Utterance}: Give the average life expectancy for countries in Africa which are \textit{republics}? \\
    \textbf{Prediction}: \texttt{SELECT Avg(LifeExpectancy) FROM country WHERE Continent = 'Africa'} \\
    \textbf{Error}: Missing \texttt{GovernmentForm = 'Republic'}. \\
    
    \midrule
    
    \textbf{Pain Point 3}: Fail to compose complex target SQL. \\
    \textbf{Utterance}: Which semesters do not have any student enrolled? List the semester name. \\
    \textbf{Prediction}: \texttt{SELECT semester\_name FROM Semesters WHERE semester\_id NOT IN (SELECT semester\_name FROM Student\_Enrolment)} \\
    \textbf{Error}: Should use \texttt{semester\_id} in nested SQL to align with the column in \texttt{WHERE} clause. \\

    \bottomrule
    \end{tabularx}%
  \caption{Error examples collected from the \textsc{Spider} development set based on the RAT-SQL + BERT~\cite{wang2019rat}.}
  \label{tab:examples}%
  \vspace{-3mm}
\end{table}%

However, based on error analysis on the output of neural language model-based text-to-SQL systems, we observe that these models can be further enhanced if we could mitigate the following three pain points, which are also illustrated in Table~\ref{tab:examples}.
(1) \emph{The model is ineffective to match and detect column names in utterances}.
The model should learn to detect column names mentioned in utterances by matching utterance tokens with the schema, and use the matched columns in the generated SQL. 
The error analysis indicates that, in some cases, models miss some columns when synthesizing the target SQL, while the column is mentioned explicitly in the utterance. 
(2) \emph{The model fails to infer the columns implicitly from cell values}.
This problem is trickier than the first one, because the model is expected to infer the column name based on some cell values mentioned in the utterance, instead of just matching the utterance tokens with the schema.
This requires the model to have more domain knowledge.
For example, as presented in the second section of Table~\ref{tab:examples}, the model should know \texttt{republics} is a \texttt{GovernmentForm}.
(3) \emph{The model should learn to compose complex queries}.
Besides the column selection, to generate a correct SQL, the model should learn to attach the selected columns to the correct clauses.
This is a non-trivial task, especially when the target SQL is complex, e.g., when the query is nested.
As shown in the last section of Table~\ref{tab:examples}, the model should learn to use corresponding column \texttt{semester\_id} in the nested SQL, instead of using column \texttt{semester\_name}.

Recent work has demonstrated that jointly pre-training on utterances and table contents (e.g., column names and cell values) can benefit downstream tasks such as table parsing and semantic parsing \cite{yin2020tabert, herzig2020tapas}.
These models are pre-trained using the Masked Language Modeling (MLM) task by either masking tokens from the \textit{utterance} input or tokens from the \textit{schema} input.
However, this learning objective can only model the alignment between the utterance and schema implicitly.
We hypothesize that, in order to cope with the three pain points previously listed, it is necessary to use pre-training objectives that enforce the learning of contextual representations that better capture the alignment between utterances and schema/table contents.

In this work, we present a language model pre-training framework, \textbf{G}eneration-\textbf{A}ugmented \textbf{P}re-training~(\textsc{GAP}),
that exploits multiple learning objectives (pre-training tasks) and synthetic data generation to jointly learn contextual representations of natural language utterances and table schema.
We propose the following three new learning objectives that not only enforce joint learning but also improve the ability of the model to grasp more domain knowledge, which is helpful in cross-domain scenarios:
(1) \emph{column prediction task},
which is a pre-training task that consists in giving a label for each column in the input schema to decide whether it is used in the input utterance or not. This task is intent to improve the column detection ability of the model.
(2) \emph{column recovery task}, 
which consists in randomly replacing some of the column names with one of their cell values and asking the model to recover the original column name either based on the cell value itself or based on the contextual information of the utterance when the column is explicitly mentioned in the utterance. This learning objective is meant to enhance the column inferring ability of the model.
(3) \emph{SQL generation},
which consists in generating SQL queries given utterances and schema.
This task can boost the ability of the model to compose complex queries by leveraging large scale SQL datasets from the Web.%, such as Github.

A key challenge to use the proposed pre-training tasks is training data.
Although it is easy to obtain large scale datasets of crawled tables and SQL queries, 
it is difficult to obtain high-quality utterances interrelated with the tables or logically consistent with crawled SQL queries.
Recent work used the surrounding text of tables as a proxy of natural language utterances~\cite{yin2020tabert, herzig2020tapas}.
However, this option is far from optimal because those texts are dissimilar to user utterances in terms of text length, composition and content.
The surrounding text of a table is usually a paragraph, while natural language utterances in the downstream task are short sentences.
Furthermore, the content of surrounding text of tables can be quite noisy because the text may be irrelevant to the table.
In \modelname,
we overcome the pre-training data challenge through the use of synthetic data.
We propose two sequence-to-sequence (seq2seq) generative models, \emph{SQL-to-text} and \emph{table-to-text}, that can produce large scale datasets with enough quality for pre-training.
We train our generative models by finetuning BART~\cite{lewis2019bart}, a state-of-the-art pre-trained language model.
Concurrently,~\citet{yu2020grappa} and~\citet{deng2020structure} utilized synthetic data generated from synchronized context-free grammar and existing data-to-text datasets~\cite{parikh2020totto} for pre-training, respectively, which requires extra crowd and expert annotation efforts.

The outcome of \modelname is a pre-trained model that can be plugged into neural semantic parsers to compute contextual representations of utterances and schema.
We apply \modelname to text-to-SQL semantic parsing datasets, and experimental results show that systems augmented with \modelname~outperform state-of-the-art semantic parsers on \textsc{Spider} and \textsc{Criteria-to-SQL} datasets.
In summary, our work presents the following main contributions:
\begin{itemize}
    \item Based on an error analysis, we spot three main issues in pre-trained LM-based text-to-SQL semantic parsers.
    \item We propose a new framework for pre-training semantic parsers that exploits multiple pre-training tasks and synthetic data.
    \item We present three novel learning objectives that alleviate the three main issues spotted with pre-trained LMs for semantic parsing. 
    \item We propose a novel strategy to overcome pre-training data challenges by leveraging SQL-to-Text and Table-to-Text generative models to generate synthetic data for learning joint representations of textual data and table schema. 
    \item To the best of our knowledge, this is the first work to effectively use both crawled SQL and crawled tables to enhance the text-to-SQL semantic parsing task. Our code is public for future work.~\footnote{\url{https://github.com/awslabs/gap-text2sql}} 
\end{itemize}

\section{Models}

\begin{figure*}[t]
    \centering
    \includegraphics[width=\textwidth]{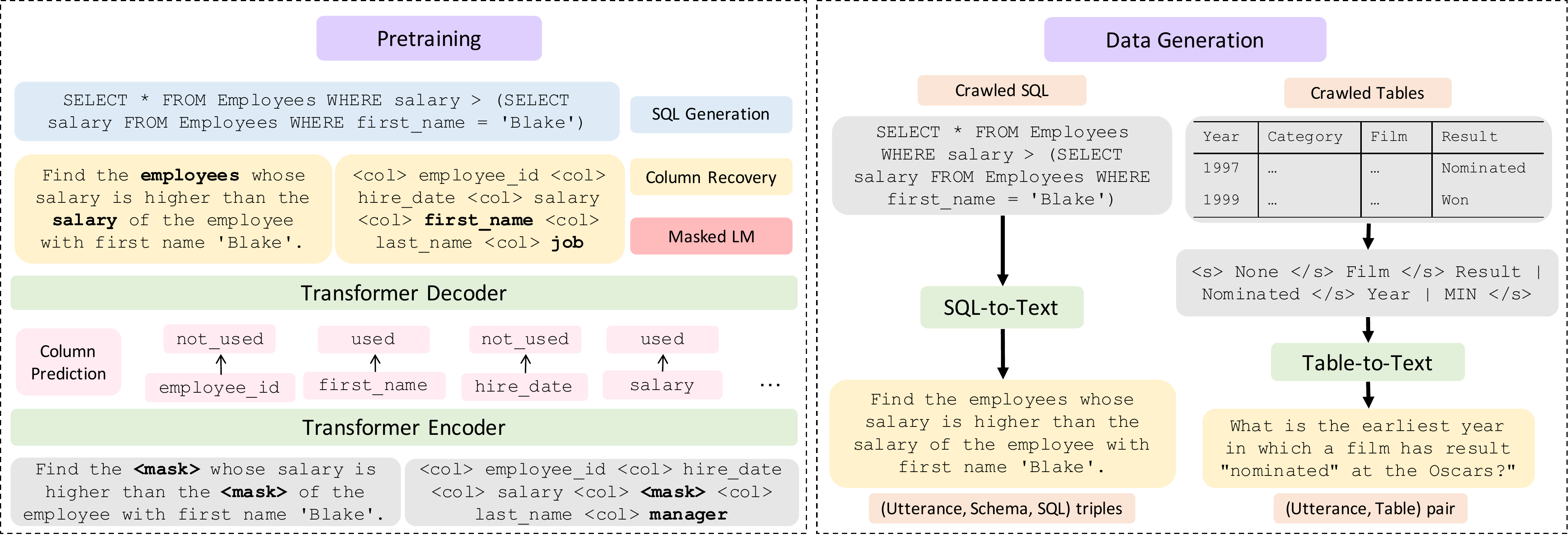}
    \caption{Building blocks of GAP framework. On the left side we illustrate our proposed pre-training tasks. On the right side we depict our proposed data generation strategies.}
    \label{fig:arch}
\end{figure*}

We first present the architecture of the semantic parsers, and then introduce the pre-training model in the \modelname framework. Lastly, we describe how to obtain the synthetic pre-training data with generative models.

\subsection{Text-to-SQL Semantic Parser}

The Text-to-SQL semantic parser translates natural language utterances to SQL queries.
The semantic parsers in our experiments are based on the encoder-decoder architecture.
Given an utterance $U$ = $\{x_1, x_2, ..., x_n\}$ and an schema $S$ consisting of tables $T = \{t_1, t_2, ..., t_{|T|}\}$ and columns $C = \{c_1, c_2, ..., c_{|C|}\}$, we leverage the contextual encoder to obtain the representations of utterance tokens and schema.
The decoder is required to compute a distribution $P(Y|X, S)$ over SQL programs.
Based on different model designs, the decoder learning target $Y$ can be raw SQL tokens~\cite{zhang19} or other intermediate representations such as SemQL~\cite{guo2019towards} or AST tree~\cite{bogin2019representing,yin2020tabert}.

\subsection{Pre-training Model}

The left part of Figure~\ref{fig:arch} presents an overview of \modelname in the pre-training stage.
Given an utterance $U$ and schema $S$, \modelnamelm~takes as input the concatenation of $U$ and the column names $c$ in $S$ in the following format $X = $ \texttt{\{<s> $U$ <col> $c_1$ <col> $c_2$ ... <col> $c_{|C|}$ </s>\}},
where $c_i$ denotes the $i$-th column in schema $S$.
With the 12-layer transformers, each token in the input can be encoded as contextual representations, denoted as $\textbf{h}$.
For different learning objectives, the representations are utilized by different decoders.
To jointly learn contextual representations for utterances and schemas and mitigate the three pain points discussed in the intro, we leverage four learning objectives in the pre-training:
Besides the Masked Language Model~(MLM), we propose learning objectives including Column Prediction~(\colpred), Column Recovery~(\colrec), and SQL Generation~(\gensql).
Multi-task learning is leveraged for these learning objectives,

\smallskip \noindent \textbf{\fullcolpred(\colpred)}:
The \fullcolpred learning objective encourages the model to capture the alignment signals between the utterance and schema, by predicting whether a column is used in the utterance or not.
An illustration is shown in the pink component of Figure~\ref{fig:arch}.
Specifically, based on the representations obtained from the transformer encoder, 
a two-layer MLP is applied on each column representation ${\textbf{g}_{col}}$, which is obtained from the output of an average pooling layer that aggregates all sub-tokens of the corresponding column.
Afterward, a sigmoid activation function is applied to obtain the probability that the corresponding column is mentioned in the utterance.
The \modelnamelm~maximizes $P_{\theta_{enc}}(Y_c|X)$ where $Y_c$ is a 0/1 label for a column and $X$ is in its unmasked version.

\smallskip \noindent \textbf{\fullcolrec(\colrec)}:
The \fullcolrec learning objective strengthens the model's ability to discover the connections between the cell values and the column names, by recovering the column name based on a sampled cell value.
For example, as shown in left yellow part of Figure~\ref{fig:arch}, the model recovers the column name \texttt{job} from cell value \texttt{manager}.
Generally, the transformer decoder recovers column names based on two information sources: 
one is the actual cell value, and
the other one is the column name mentions in the utterance. 
We design the following rules for the value replacement:
\begin{itemize}
\item If a column is not mentioned in the utterance, we will replace the column name with its cell value with probability of 0.5.
In this case, the column name will be recovered from cell value without other contextual information.
\item If a column is mentioned, we will directly replace the column name with its cell value.
In this case, the model can leverage the contextual information from the utterance and the cell value to recover the column name.
\end{itemize}

\smallskip \noindent \textbf{\fullgensql(\gensql)}:
This learning objective is directly related to the downstream task.
Based on the representation from the transformer encoder, the \modelnamelm~decoder maximizes $p_{dec}(y_{sql}|\textbf{h})$.
This learning target encourages the model to learn to compose complex SQL that requires logical reasoning, considering that there are a large number of sophisticated SQLs in crawled data.
For example, the \modelnamelm~decoder needs to generate the column in appropriate position such as in the \texttt{ORDER BY} clause or \texttt{WHERE} clause, instead of just predicting the column is used or not.
Specifically, the \modelnamelm~decoder emits the target SQL token by token with a close vocabulary set, which is composed of the SQL keywords vocabulary and column names.
The embeddings of the SQL keywords are randomly initialized and trained during the pre-training phase.
The column representations are obtained in the same way as the one used in \fullcolpred learning objective, by averaging the column's sub-tokens representations.
At each decoding step, the decoder generates a hidden vector and then a dot-product operation is applied on it and the target vocabulary representations, yielding a probability distribution over the vocabulary set.

\smallskip \noindent \textbf{Masked Language Model(MLM)}: 
We use the standard MLM objective, with a masking rate of 35\% sub-tokens in the whole input sequence, including the utterance and schema.
Based on the representation from transformer encoder, \modelnamelm~employs a transformer decoder to maximize $p_{\theta}(x | x_m)$ on large scale utterance-schema pairs, where $x_m$ is the masked version of $x$.

\subsection{Pre-training Data Generation}

As discussed, previous pre-training approaches such as TaBERT~\cite{yin2020tabert} and TAPAS~\cite{herzig2020tapas} use the surrounding texts of the tables as a proxy of natural language utterance.
However, those texts are noisy and sometimes are not directly related to the table contents.
In the downstream task, the input texts are usually utterances/user queries, which are short and highly dependent on the schema and contents of the structured data.
In order to minimize the gap between pre-training and downstream tasks, we adopt a state-of-the-art pre-trained sequence-to-sequence model, such as BART, to generate high-quality utterances based on crawled SQLs or structured tables.

As shown in the right part of Figure~\ref{fig:arch}, we design two different models, namely SQL-to-Text generation model and Table-to-Text generation model, for handling the two different inputs.
Specifically, the SQL-to-text generation model takes the SQL as input and generates the utterance that explains the query intent.
The other model, the Table-to-Text generation model, generates utterances based on a set of sampled column names and cell values from tables.
In this way, we can generate utterances interrelated with tables without composing queries that might be suspicious.

\smallskip \noindent \textbf{SQL-to-Text Generation:}
We crawl 30K SQLs from GitHub\footnote{\url{https://github.com}}.
To generate utterances for these SQL queries, we train a SQL-to-Text model on the \textsc{Spider} dataset.
The input is the original SQL and it is directly tokenized by the BART tokenizer without additional pre-processing.
After finetuning BART, the model can generate high-quality utterances logically consistent with the input SQL,
achieving a 0.1934 BLEU score on the development set.
Then we use the model to generate utterances for crawled SQLs.
We further extract columns and tables in each SQL as positive schema candidates, denoted as $\texttt{schema}_{pos}$. 
We also sample columns and tables from the pool which are extracted from other SQLs as negative candidates, denoted as $\texttt{schema}_{neg}$.
The final schema is composed of these two parts.
The utterance-schema-SQL triples are then collected for the \gensql~learning objective in the pre-training phase.

\smallskip \noindent \textbf{Table-to-Text Generation:}
Generating utterances from tables is different because query intents are not given.
Instead of synthesizing noisy SQLs and then translating into natural language utterances, we propose a Table-to-Text generation model that can directly transform a set of column names and cell values into user queries without query intent constrains.
Specifically, we sample column names and cell values~(both are referred as candidates) from tables. 
For example, based on the table in the right part of Figure~\ref{fig:arch}, we can sample columns \texttt{Year}, \texttt{Film} and \texttt{Result}, and a cell value \texttt{Nominated}.
We then linearize the sampled candidates into \texttt{\{column name | associated cell value list\}} and concatenate them into a sequence, separated by \texttt{<sep>} token.
Furthermore, to control the complexity and diversity of the generated text, we integrate three types of control codes into the model input:
\begin{itemize}
\item Aggregator-based control code: Including \texttt{COUNT}, \texttt{MAX}, \texttt{MIN}, \texttt{AVG}, and \texttt{SUM}.
For the first two sampled columns, we randomly sample an aggregator for each with the probability $\gamma_1$~(we use $\gamma_1$ as 0.5) 
if the column type matches with the selected aggregator, e.g., aggregator \texttt{SUM} should be applied on numerical type column.
If the control codes are sampled, they will be appended to the associated cell value list of the corresponding column.
\item Structure control code: Including \texttt{IN}, \texttt{NOT IN}, \texttt{INTERSECT}, \texttt{UNION}, and \texttt{EXCEPT}. 
For each example, with probability of $\gamma_2$~(we use $\gamma_2$ as 0.35), we randomly sample one of them with uniform distribution. Otherwise, \texttt{NONE} is used.
This control code is used as the first item of the input sequence.
\item Order-based control code: We add \{\texttt{LIMIT} : number\} as a part of the control code, which is usually used in an \texttt{ORDER BY} based query.
With this control code, the generated utterances usually contain phrases that constrain the number of query results should be returned, e.g., \textit{Show the name of aircrafts with top three lowest speed.}.

\end{itemize}

We fine-tune a BART model on \textsc{Spider} dataset to create the generator.
To align with our designed input, we convert the SQL into the format we expected.
We extract all the columns and their associated aggregators and values from the SQL. 
We also obtain any special control codes that appears in the SQL.
After fine-tuning, the model achieves 0.1821 BLEU score on the development set.
Afterwards, we apply the finetuned model to the crawled tables and generate high-quality utterances.
The utterance-schema pairs are collected for the learning objectives including MLM, \colpred, and \colrec~in pre-training phase.

For the pre-training step, we need to decide whether a column is mentioned in the utterance or not.
To create the label for this, we directly regard all the sampled columns to have positive label.
This is based on the assumption that the generation model uses all the columns to synthesize the utterance, and does not have the hallucination issue that models generate some columns names or cell values that are not presented in the input.

\section{Experiments}

\subsection{Configurations}
In the pre-training, we train our \modelnamelm~with the underlying transformers initialized with BART~\cite{lewis2019bart} model.
During the fine-tuning phase, we only leverage the encoder component of the \modelnamelm~with 12-layer transformers as the contextual encoder for the semantic parsers.
For more details, please refer to appendix.

\subsection{Tasks, Datasets and Baseline Systems}

For the downstream tasks, we conduct experiments on two datasets to show the effectiveness of our framework.

\smallskip \noindent \textbf{\textsc{Spider}}: \textsc{Spider} dataset~\cite{yu2018spider} is a text-to-SQL dataset with 10,181 annotated parallel utterance-database-SQL triples.
Different from WikiSQL, the examples in the \textsc{Spider} dataset is more complex, 
involving nested query, set operation, and multiple tables joining.
The exact set match accuracy is the evaluation metrics.
The test set is not publicly available.
For baseline parser, we use RAT-SQL~\cite{wang2019rat} model as our baseline system to report the end-to-end performance.
RAT-SQL model is the state-of-the-art parser in the \textsc{Spider} dataset, which leverages the 8-layer relation-aware transformer to model the connections among tables and utterance.
To show that the \modelnamelm~can be plugged into different neural semantic parsers, we further use IRNet~\cite{guo2019towards} model for ablation study.
IRNet semantic parser is based on SemQL grammar, which is an effective intermediate representation for SQL.
IRNet is efficient in terms of training time, which requires 1 day for training, while RAT-SQL model requires approximately 5 days for training.
We augment the encoder part of our \modelnamelm~to these base parsers, by replacing their original contextual encoders.

\smallskip \noindent \textbf{\textsc{Criteria-to-SQL}}: \textsc{Criteria-to-SQL} is a dataset to facilitate retrieving eligible patients for a trial from the electronic health record database.
The task is to translate the eligibility criteria to executable SQL queries.
For example, a criteria statement \textit{any infection requiring parenteral antibiotic therapy or causing fever (i.e., temperature $>$ 100.5f ) $\leq$ 7 days prior to registration} is required to be interpreted into SQL \texttt{SELECT id FROM records WHERE active\_infection = 1 AND (parenteral\_antibiotic\_therapy = 1 OR causing\_fever = 1 OR temperature $>$ 100.5)}.
The dataset contains 2003 annotated examples, and the evaluation metrics are the SQL accuracy and execution accuracy.
Our baseline system for \textsc{Criteria-to-SQL} dataset is adopted from \cite{yu2020dataset}, a slot-filling based model that takes advantages of the prior grammar knowledge to design the sketch.
We denote this system as \textbf{YXJ} model.
The system uses the BERT-base as the contextual encoder.

\subsection{Spider Results}

\begin{table}[t]
  \centering
  \small
    \begin{tabular}{lcc}
    \toprule
    Model & Dev & Test \\
    \midrule
    Global-GNN (Bogin et al., 2019a) & 0.527 & 0.474 \\
    EditSQL + BERT~\cite{zhang19} & 0.576 & 0.534  \\
    IRNet + BERT~\cite{guo2019towards} & 0.619 & 0.547  \\
    RyanSQL V2 + BERT~\cite{Choi2020RYANSQLRA} & 0.706 & 0.606 \\
    RAT-SQL V2 + BERT~\cite{wang2019rat} & 0.658 & 0.619 \\
    AuxNet + BART & 0.700 & 0.619 \\
    ShadowGNN$^{\dag}$ & - & 64.8 \\
    YCSQL + BERT$^{\dag}$ & - & 65.3 \\
    RAT-SQL V3 + BERT~\cite{wang2019rat} & 0.697 & 0.656  \\
    RAT-SQL + STRUG~\cite{deng2020structure} & 0.727 & - \\
    RAT-SQL + GraPPa~\cite{yu2020grappa}$^{\dag}$ & \textbf{0.734} & 0.696 \\
    \midrule
    RAT-SQL + BERT~(our replicate) & 0.665 & -  \\
    RAT-SQL + BART Encoder~(ours) & 0.676 & 0.651  \\
    RAT-SQL + \modelnamelm~(ours) & 0.718 & \textbf{0.697}\\
    \bottomrule
    \end{tabular}%
  \caption{Exact set match accuracy on the public development set and hidden test set of \textsc{Spider}. {\dag} denotes that the algorithms are concurrent work and leaderboard results are public after our paper submission.}
  \label{tab:spider}%
\end{table}%

Table~\ref{tab:spider} shows the end-to-end results on \textsc{Spider} dataset.
Based on the codebase\footnote{\url{https://github.com/microsoft/rat-sql}} provided by \citet{wang2019rat}, we replicate the RAT-SQL + BERT large model, achieving 0.665 exact set match accuracy on the development set.
This matches the RAT-SQL V2 + BERT but still worse than its V3.
By replacing the BERT-large with the encoder of BART\footnote{The encoder of BART has 12-layer transformers while BERT-large has 24-layer transformers.}, we obtain accuracy of 0.676 on the development set and 0.651 on test set.
The BART Encoder based model achieves comparable results with RAT-SQL V3 + BERT large model on the hidden test set with less encoder layer~(BART encoder has 12-layer transformers while BERT large model has 24-layer transformers).
With our \modelnamelm, the RAT-SQL can be further augmented, benefiting from enhanced contextual encoding ability.
The model achieves accuracy of 0.718 on the development set and 0.697 on the hidden test set.
This confirms the effectiveness of the Generation-augmented pre-training.
This performance achieves the state-of-the-art performance on the hidden test set with less model parameters on \textsc{Spider} dataset at the time of writing.
Comparing scores of the development set and the test set, we observe BART based models~(+BARR Encoder or \modelnamelm) have better generalization ability on the hidden test, considering that the gap between the development set and test set is smaller than the model such as RAT-SQL V3 + BERT.
Concurrently,~\citet{yu2020grappa} used synchronized context-free grammar to generate synthetic data for pre-training;~\citet{deng2020structure} leveraged existing large-scale data-to-text dataset for enhancing the structured data representations. Both of them achieve comparable performance as ours, but require more model parameters~(24-layer transformers in the pre-trained model) and extra crowd and expert annotation efforts.

\begin{table}[t]
  \centering
  \small
    \begin{tabular}{lccccc}
    \toprule
    RAT-SQL & \multicolumn{1}{l}{Easy} & \multicolumn{1}{l}{Medium} & \multicolumn{1}{l}{Hard} & \multicolumn{1}{l}{Extra} & \multicolumn{1}{l}{All} \\
    \midrule
     +BERT & 0.830 & 0.713 & 0.583 & 0.384 & 0.656 \\
     +BART Encoder & 0.826 & 0.711 & 0.581 & 0.370 & 0.651 \\
     +\modelnamelm & 0.872 & 0.751 & 0.637 & 0.412 & 0.697 \\
    \bottomrule
    \end{tabular}%
  \caption{Breakdown results on \textsc{Spider} hidden test set.} 
  \label{tab:breakdown}%
\end{table}%

Based on the complexity of the SQL, the examples in \textsc{Spider} are classified into four types: Easy, Medium, Hard, Extra Hard.
Here, we provide a breakdown analysis on the \textsc{Spider} test set, as shown in Table~\ref{tab:breakdown}.
The BERT results are adopted from \citet{wang2019rat}, which is the state-of-the-art system on \textsc{Spider} dataset.
Comparing the RAT-SQL+BERT model and RAT-SQL+BART Encoder model, we can find that the performance of RAT-SQL+BART is comparable with the state of the art, but with fewer model parameters~(12-layer transformers in BART encoder v.s. 24-layer transformers in BERT-large encoder).
We also find that the RAT-SQL+\modelnamelm~Encoder can have significant improvement over its baseline RAT-SQL+BART Encoder on each hardness level.

\begin{table}[t]
  \centering
  \small
    \begin{tabular}{lccc}
    \toprule
    RAT-SQL & \multicolumn{1}{l}{Selection} & \multicolumn{1}{l}{Inferring} & \multicolumn{1}{l}{Composing} \\
    \midrule
     +BART Encoder & 14 & 10 & 16 \\
     +\modelnamelm & 5 & 6 & 7  \\
    \bottomrule
    \end{tabular}%
  \caption{Error counts of different types for RAT-SQL+BART Encoder and RAT-SQL+\modelnamelm~Encoder.} 
  \label{tab:errorstat}%
\end{table}%

For comparison, we sample 40 examples from \textsc{Spider} development set which the baseline system RAT-SQL+BART Encoder fails in.
Because we focus more on the following three error types as we discussed in the introduction part: \textit{column selection error}, \textit{column inferring error} and \textit{SQL composing error}, we ignore other error types during the sampling.
We analyze the predictions of both the RAT-SQL+BART Encoder and RAT-SQL+\modelnamelm~Encoder.
The statistics are shown in Table~\ref{tab:errorstat}.
The numbers in the Table represent the error count of each error type.
Based on the analysis results, we can find that the \modelnamelm~Encoder can alleviate all the three error types, especially the \textit{column selection} and \textit{SQL composing error}.

\subsection{Criteria-to-SQL Results}

\begin{table}[t]
  \centering
  \small
    \begin{tabular}{lcc}
    \toprule
    Model & \multicolumn{1}{l}{SQL Acc.} & \multicolumn{1}{l}{Exec. Acc.} \\
    \midrule
    SQLNet~(Xu et al. 2017) & 0.132 & 0.139 \\
    % SQLova~\cite{hwang2019comprehensive} & 0.119 & 0.129 \\
    YXJ~\cite{yu2020dataset} & 0.142 & 0.158 \\
    \midrule
    YXJ + Roberta~(ours) & 0.294 & 0.538 \\
    YXJ + BART Encoder~(ours) & 0.307 & 0.558 \\
    YXJ + \modelnamelm~(ours) & \textbf{0.327} & \textbf{0.594} \\
    \bottomrule
    \end{tabular}%
    \caption{Test results of Criteria-to-SQL. The SQL accuracy and the execution accuracy are reported.}
  \label{tab:criteria-to-sql}%
  \vspace{-0.1cm}
\end{table}%

Table~\ref{tab:criteria-to-sql} shows the test results of the \textsc{Criteria-to-SQL} dataset.
The YXJ model~\cite{yu2020dataset} is built upon BERT-base encoder
and sketch-based decoder, achieving the state-of-the-art performance of 0.142 SQL accuracy and 0.158 execution accuracy.
We use this system as our baseline.
Instead of using the BERT encoder, we augment the model with more powerful pre-trained language models such as RoBERTa and BART.
These two pre-trained language models yield significant improvement over the BERT baseline, achieving 0.294 and 0.307 on the SQL accuracy, respectively.
After executing the generated SQL queries against the database, these two models obtain 0.538 and 0.558 execution accuracy, respectively.
By replacing the BART encoder with \modelnamelm, the parser obtains 2.0\% improvement on the SQL accuracy and 3.6\% improvement on the execution accuracy, which registers new state-of-the-art performance.
This also confirms our assumption that the parsers can benefit from better quality of contextual encoders that jointly reason over utterances and schemas.

\subsection{Impact of Learning Objectives}

\begin{table}[t]
  \centering
  \small

    \begin{tabular}{l|cc}
    \toprule
    \multicolumn{1}{l}{Model} & \multicolumn{2}{c}{Dev. Acc.} \\
    \midrule
    IRNet + BERT (Ours) & \multicolumn{2}{c}{0.620} \\
    IRNet + TaBERT & \multicolumn{2}{c}{0.652} \\
    IRNet + RoBERTa (Ours) & \multicolumn{2}{c}{0.658} \\
    \midrule \midrule
    Learning Objectives & \multicolumn{1}{l}{w/o GenSQL} & \multicolumn{1}{l}{w/ GenSQL} \\
    \midrule
    baseline & 0.680 & 0.699 \\
    \midrule
    MLM   & 0.697 & 0.717 \\
    \colpred & 0.699 & 0.710 \\
    \colrec &  0.705 & 0.719 \\
    MLM + \colpred & 0.704 & 0.716 \\
    MLM + \colrec & 0.711 &  0.728\\
    MLM + \colpred + \colrec & 0.715 & 0.723 \\
    \bottomrule
    \end{tabular}%
    \caption{Ablation study on different learning objectives.}
  \label{tab:ablation:objectives}%
  \vspace{-0.3cm}
\end{table}%

We investigate four different learning objectives in this work, namely Masked Language Model~(MLM), \fullcolpred(\colpred), \fullcolrec(\colrec) and \fullgensql(\gensql).
We conduct the ablation study on \textsc{Spider} development set to compare the first three learning objectives under two different conditions: One is with \gensql~learning objective and the other one is without. 
We use the IRNet based model in the ablation study because it is more efficient in training than RAT-SQL based model, and it can achieve comparable performance.
We also want to show that our \modelnamelm~is plugin-able and can augment different semantic parsers.
Table~\ref{tab:ablation:objectives} shows the ablation results.

The first section of the Table~\ref{tab:ablation:objectives} shows the results of three baseline systems that are based on IRNet model: IRNet + BERT, IRNet + TaBERT and IRNet + RoBERTa.
These results confirm that improving the encoder quality of the semantic parser is a promising direction to pursue.

In the second section of the Table~\ref{tab:ablation:objectives}, we present detailed ablation study results. Without the \gensql~learning objective, compared with baseline~(IRNet + BART Encoder), the three learning objectives~(MLM, \colpred, \colrec) can improve the performance of the parser, with a 1.7\%, 1.9\% and 2.5\% increase, respectively.
This indicates that these learning objectives improve the encoding quality of the transformer encoder.
Based on the standard unsupervised learning objective MLM, we observe that the \colpred~and \colrec~learning objectives are helpful, 
which lead the model to the accuracy of 0.704 and 0.711, respectively.
When we further combine the three learning objectives, the semantic parser's effectiveness is furthered boosted, achieving accuracy of 0.715, a 3.5\% increase over its baseline.

With the \gensql~learning objective, the comparison of these three learning objectives is based on a higher baseline with accuracy of 0.699.
This indicates that the \gensql~learning objective is valuable.
Under this experimental condition, we observe that the MLM learning objective brings consistent improvement over the baseline with 1.8\% increase on the accuracy.
For the \colpred~and \colrec, the accuracy is boosted by 1.1\% and 2.0\%, respectively.
When we combine the MLM with the \colpred, we only observe comparable results with the MLM, without further significant improvement.
However, the \colrec~learning objective brings the MLM a step forward, achieving the 0.728 on the accuracy.
The combination of the three learning objectives under w/ \gensql~condition improve 2.4\% on accuracy over the baseline.
These results show that \gensql and \colrec are two salient learning objectives, leading the model to obtain accuracy more than 0.720, registering a new state-of-the-art performance on public development set on \textsc{Spider}.

\subsection{Analysis of Pre-Training Inputs}

\begin{table}[t]
  \centering
  \small
    \begin{tabular}{lc}
    \toprule
    Learning Objective & Dev. Acc. \\
    \midrule 
    baseline & 0.680 \\
    \midrule
    MLM   & 0.697 \\
    MLM w/o utterance & 0.678~(-1.7\%) \\
    MLM w/o schema & 0.679~(-1.6\%) \\
    MLM (surrounding text) & 0.679~(-1.6\%) \\
    \midrule
    CRev & 0.705 \\
    CRev w/o utterance & 0.688~(-1.7\%) \\
    CRev (surrounding text) & 0.697~(-0.8\%) \\
    \bottomrule
    \end{tabular}%
  \caption{The ablation study on different inputs for the pre-training based on the IRNet based model.}
  \label{tab:ablation:nlsentence}%
  \vspace{-0.3cm}
\end{table}%

\smallskip \noindent \textbf{Whether to use utterance in pre-training}: To prove that the utterance is beneficial in the pre-training,
we conduct an ablation study by comparing the pre-trained models which are trained with and without utterance.
Our experiments are based on the MLM and \colrec~learning objectives because the other two~(\colpred~and \gensql) require the utterance as the input based on their task definitions.
Similarly, we use IRNet as our base parser.

The experimental results on \textsc{Spider} development set are shown in Table~\ref{tab:ablation:nlsentence}.
As we can see, if the \modelnamelm~is trained with MLM learning objective without utterances as part of the input, the semantic parser performance drops to 0.678 from 0.697, which is lower than the baseline~(0.680) by 0.2\%.
For the \colrec~learning objective, the accuracy drops from 0.705 to 0.688, a 1.7\% decrease, if the \modelname is trained without utterance.
Even though, \colrec~learning objective trained without utterances is still helpful, which improves the baseline model by 0.8\%.
This aligns with our analysis of the \colrec~learning objective: model can leverage two information sources to recover the column name.
If there are no utterances, the model can only use the signals the cell values provide to recover the column name.
Furthermore, when the model can access more contextual information, which is provided by the utterance, the model can learn better encoding ability by learning to align the cell values and the column names in the utterances.

\smallskip \noindent \textbf{Whether to use schema in pre-training}: Another input choice is to only keep the utterances in the pre-training.
This experimental setting is to justify that the model's improvement is not solely from better utterance representation.
This input strategy is only applicable to the MLM learning objective as the schema is a necessary component for other learning objectives.
As shown in the MLM w/o schema entry in Table~\ref{tab:ablation:nlsentence}, the model performance drops to 0.679, indicating that learning joint utterance and schema representation is necessary for this task.

\smallskip \noindent \textbf{Whether to use the generated text or the surrounding text of the table}: 
The value of the generated text is already justified by the learning objectives such as \colpred~or \gensql, because the definitions of these learning objectives require the generated utterances that cannot be obtained from the surrounding text of the table~(denoted as surrounding text).
Here, we further rationalize our generation-augmented framework on MLM and \colrec~learning objectives by replacing the generated text with the surrounding text.

The results are presented in the entries of MLM (surrounding text) and \colrec 
~(surrounding text) of Table~\ref{tab:ablation:nlsentence}.
Overall, we can observe that the generation technique is superior to using the surrounding text as a proxy in the MLM and \colrec~ learning objectives, considering the models drop 1.6\% and 0.8\% on accuracy, respectively.
We also find that the \colrec learning objective is more robust for pre-training, given that the fine-tuned model performance gets less influence compared with the one with MLM learning objective.

\subsection{Analysis of Pre-trained Model}

As the \modelnamelm~provides gains on the text-to-SQL benchmarks, understanding what they learn is important.
Following previous work~\cite{liu2019linguistic, hewitt2019designing, hewitt2019structural},
we design a probing task, Column-Value Matching~(CVM), 
to examine the extent to which the model can align the cell values in the utterances and the columns, i.e., the probes need to predict which column the cell value belongs to.

Specifically, given the column spans and cell value spans~(part of utterances), we can obtain their representations with contextual encoders such as BART or \modelnamelm~Encoder, and an average pooling layer.
We further compress the representations into another space with linear transformation, denoted as $\{\textbf{c}_j\}$ and $\textbf{v}_i$, respectively. 
The probability of selecting column $c_j$ given cell value $v_i$ is determined by $p(\textbf{c}_j | \textbf{v}_i) \propto exp(\textbf{v}_i \textbf{c}_j)$.
During training, parameters of language model encoders are fixed.
Here, we conduct the probing task training on the \textsc{Spider} dataset.

Note that the unavailability of span annotations of cell values in \textsc{Spider} dataset leads to further data pre-processing.
Since human annotation is costly, we try to annotate the spans by automatic aligning the cell values in SQL to the utterance tokens.
For a cell value used in SQL, assuming it has $n$ tokens, we obtain all n-grams from the utterance, and select the best candidate based on the fuzzy matching score\footnote{\url{https://github.com/seatgeek/fuzzywuzzy}}~(determined by Levenshtein Distance) when the score is higher than a threshold~(we use 60 in our experiment). 
For integers in the SQL, we also leverage a dictionary to map it to English words when searching for their matches.
If n-gram candidates are founded, the cell value will be used in the probing experiment.
During the training, the encoder~(e.g. BART Encoder or \modelnamelm~Encoder) is fixed and only the parameters of probes are tune.
The probes are optimized with Adam optimizer with cross-entropy loss function.
The learning rate is $1e-5$ and the model is trained for 100 epochs on \textsc{Spider} dataset with batch size of 96.
The evaluation metric is instance-level accuracy, i.e., the prediction is correct if every cell value used in the utterance is matched with the correct column.

\begin{table}[t]
  \centering
  \small
    \begin{tabular}{lc}
    \toprule
    Model & \multicolumn{1}{l}{Match Acc.} \\
    \midrule
    BART Encoder   & 23.17 \\
    \midrule
    \modelnamelm~(MLM) Encoder &  32.72 \\
    \modelnamelm~(MLM + CRec) Encoder & 36.78 \\
    \modelnamelm~(MLM + CPred) Encoder & 44.51 \\
    \bottomrule
    \end{tabular}%
    \caption{Results of Value-Column Matching Probing Task.}
  \label{tab:analysis:colprobing}%
  \vspace{-0.3cm}
\end{table}%

The results are shown in Table~\ref{tab:analysis:colprobing}.
We report the accuracy of the BART Encoder model as our probing baseline, which achieves accuracy of 23.17\%.
With \modelnamelm~(MLM) Encoder, the accuracy raises to 32.72\%, indicating that the model learns to align the cell values and column names implicitly.
By providing stronger supervisions, the MLM+\colrec~based model and MLM+\colpred~based models obtain higher accuracy~(36.78\% and 44.51\%),
showing that the models capture more alignment signals, contributing to better semantic parser performance.

\section{Related Work}
\label{related_work}

\smallskip \noindent \textbf{Semantic Parsing}:
The semantic parsing task is framed as mapping the natural language utterances to meaning representations.
The meaning representations can be executed in a variety of environments such as data analysis by translating the natural language queries into database queries.
Based on different meaning representations, the semantic parsing task can be classified into three regimes~\cite{kamath2018survey}: logic based formalism such as $\lambda$-DCS~\cite{liang2013lambda}, graph based formalism such as AMR~\cite{banarescu2013abstract} and UCCA~\cite{abend2013universal}, and programming languages such as Python and SQL. 
Recently, more interests are concentrated on the SQL-based semantic parsing,
and most of the work try to solve the problem with general encoder-decoder architecture.
Overall, they enhance the models based on following aspects: 
(1) Improving the decoding mechanism~\cite{yin2017syntactic,dong2018coarse,rubin2020smbop};
(2) Improving the decoding target~\cite{guo2019towards};
(3) Improving the model encoding ability~\cite{wang2019rat, bogin2019global, yin2020tabert, scholak2020duorat, ma2020mention, deng2020structure, yu2020grappa};
(4) Reranking over the generated candidates to improve parses quality~\cite{kelkar2020bertrand, yin2019reranking}.
\modelname advances the line of (3) by leveraging generation models and three novel learning objectives to enhance the utterance-schema representations.

\smallskip \noindent \textbf{Question Generation and Table-to-Text Generation}:
The question generation task is to generate grammatically and semantically correct questions.
The generated questions are usually used for enhancing the question answering models~\cite{duan2017question, guo2018question, yu2020generating, zhong2020grounded}. 
The table-to-text generation task is to generate declarative sentences that describe the information provided by the table~\cite{liu2017table,gong2019table,parikh2020totto, radev2020dart}.
Our Table-to-Text model is a combination of these two directions, focusing on generating questions from table, i.e., composing questions based on the sampled columns and cell values, without providing the detailed information about ``what to ask''.

\smallskip \noindent \textbf{Pre-training Models}:
Recent pre-training techniques exploit external knowledge~(e.g. entity-level information, commonsense knowledge, knowledge graph) into large-scale pretrained language models~\cite{xiong2019pretrained,wang2020k,peters2019knowledge,rosset2020knowledge}. 
More recently, \citet{yin2020tabert}, \citet{herzig2020tapas}, leverage the semi-structured table data to enhance the representation ability of language models.
Concurrently, \citet{yu2020grappa} and \citet{deng2020structure} leveraged synchronous context-free grammar to generate synthetic data and utilized existing high-quality data-to-text dataset for pre-training, respectively.
Different from these work, we explore the direction of utilizing the generators to enhance the joint utterances and structured schema encoding ability of the pre-trained models.

\section{Conclusion}
In this work, we spot three pain points in the Text-to-SQL semantic parsing task, 
and propose a generation-augmented pre-training framework to alleviate them, with four different learning objectives.
Experimental results on \textsc{Spider} dataset and \textsc{Criteria-to-SQL} dataset show the effectiveness of this framework, which achieves state-of-the-art performance on both datasets.

\bibliography{aaai21}

\clearpage
\appendix

\section{Appendices}

\subsection{Pre-training Data}

\smallskip \noindent \textbf{Utterance-Table Pairs}:
We extract the tables from English Wikipedia.
We further apply the following heuristic strategies to pre-process the extracted tables:
(1) Removing tables with less than 4 columns;
(2) Removing tables with less than 3 rows;
(3) Removing columns whose names have more than 10 tokens;
(4) Removing columns whose cell values have more than 50\% empty string;
(5) Filtering cell values with more than 5 tokens or contains any pre-defined non-ASCII characters.
After the pre-processing, we obtain 500K tables.

For each table, we then randomly sample the column names, cell values and control codes as the Table-to-Text generation model input to produce the utterances.
We apply the following strategies to sample inputs:
(1) We randomly generate a integer from 2 to 6, denoting the number of columns we will sample;
(2) We sample the wildcard $*$ with probability of 0.2;
(3) We sample one of the structure control codes with probability of 0.35;
(4) We sample the order-based control code with probability of 0.25;
(5) For the first two sampled columns, we randomly sample one of the aggregators with probability of 0.5;
(6) For each column without any associated aggregator-based control code, we sample one value from that column with probability of 0.4.
We then generate 4 instances per table and we finally obtain 2M training instances.

\smallskip \noindent \textbf{Utterance-Schema-SQL Triples}:
We crawl the SQL from GitHub repositories if the SQL can be parsed by one of the SQL parsers: moz-sql-parser\footnote{\url{https://github.com/mozilla/moz-sql-parser}} and sqlparse\footnote{\url{https://github.com/andialbrecht/sqlparse}}.
We apply the trained SQL-to-Text generation model to the SQL and obtain 30K utterance-SQL pairs.
To obtain the schema, for each SQL, we extract the table names and column names from the SQL as positive candidates and randomly sample table names and column names from other SQL as negative candidates.
The combination of these two components are regarded as the associated schema.
We then obtain utterance-schema-SQL triples for \gensql~learning objective training.

\subsection{Training Details}

As discussed in the previous section, each epoch contains 2M utterance-table pairs and 30K utterance-schema-SQL triples.
We train the \modelnamelm~with multi-task training strategies:
30K utterance-schema-SQL triples are for \gensql~learning objective and 2M utterance-table pairs are evenly split for the other three learning objectives, including MLM, \colpred~and \colrec.
We train the model for 6 epochs with batch size of 64 on 4 Tesla V100 GPUs.
The model is optimized with Adam optimizer~\cite{kingma2014adam} with a learning rate of $1e-5$ and linearly decayed learning rate schedule.

\subsection{Analysis for \textsc{Spider} Dataset}

\begin{table*}[t]
  \centering \small
    \begin{tabularx}{\linewidth}{L{0.97 \linewidth}}
    \toprule
    \textit{Example 1} \\
    \textbf{Utterance}: What type of pet is the youngest animal, and how much does it weigh? \\
    \textbf{Baseline}: \texttt{SELECT PetType, pet\_age FROM Pets ORDER BY pet\_age ASC LIMIT 1} \\
    \textbf{Our Model}: \texttt{SELECT PetType ,  weight FROM pets ORDER BY pet\_age ASC LIMIT 1} \\
    \midrule 
    \textit{Example 2} \\
    \textbf{Utterance}: What is the average and maximum age for each pet type? \\
    \textbf{Baseline}: \texttt{SELECT T2.PetType, Avg(T3.Age), Max(T2.pet\_age) FROM Has\_Pet AS T1 JOIN Pets AS T2 ON T1.pet\_id = T2.pet\_id JOIN Student AS T3 ON T1.student\_id = T3.student\_id GROUP BY T2.PetType} \\
    \textbf{Our Model}: \texttt{SELECT avg(pet\_age) ,  max(pet\_age) ,  pettype FROM pets GROUP BY pettype} \\
    \midrule \midrule
    \textit{Example 3} \\
    \textbf{Utterance}: What are the names of the singers who performed in a concert in 2014? \\
    \textbf{Baseline}: \texttt{SELECT T2.Name FROM singer\_in\_concert AS T1 JOIN singer AS T2 ON T1.singer\_id = T2.singer\_id JOIN concert AS T3 ON T1.convert\_id = T3.concert\_id WHERE T1.concert\_ID = 2014} \\
    \textbf{Our Model}: \texttt{SELECT T2.name FROM singer\_in\_concert AS T1 JOIN singer AS T2 ON T1.singer\_id  =  T2.singer\_id JOIN concert AS T3 ON T1.concert\_id  =  T3.concert\_id WHERE T3.year  =  2014} \\
    \midrule 
    \textit{Example 4} \\
    \textbf{Utterance}: Count the number of United Airlines flights arriving in ASY Airport. \\
    \textbf{Baseline}: \texttt{SELECT Count(*) FROM airlines AS T1 JOIN flights AS T2 ON T2.Airline = T1.uid WHERE T1.Airline = 'United Airlines' AND T2.Airline = 'ASY'} \\
    \textbf{Our Model}: \texttt{SELECT count(*) FROM AIRLINES AS T1 JOIN FLIGHTS AS T2 ON T2.Airline  =  T1.uid WHERE T1.Airline  =  'United Airlines' AND T2.DestAirport  =  'ASY'} \\
    \midrule \midrule
    \textit{Example 5} \\
    \textbf{Utterance}: What are the different continents and the total popuation and average life expectancy corresponding to each, for continents that have an average life expectancy less than 72? \\
    \textbf{Baseline}: \texttt{SELECT Count(*), Avg(LifeExpectancy), Avg(LifeExpectancy) FROM country WHERE LifeExpectancy < 72 GROUP BY country.Continent} \\
    \textbf{Our Model}: \texttt{SELECT sum(Population) ,  avg(LifeExpectancy) ,  Continent FROM country GROUP BY Continent HAVING avg(LifeExpectancy)  <  72} \\
    \midrule
    \textit{Example 6} \\
    \textbf{Utterance}: Give the ids of documents that have between one and two paragraphs. \\
    \textbf{Baseline}: \texttt{SELECT T2.Document\_ID FROM Paragraphs AS T1 JOIN Documents AS T2 ON T1.Document\_ID = T2.Document\_ID GROUP BY T1.Document\_ID HAVING Count(*) < 2} \\
    \textbf{Our Model}: \texttt{SELECT Document\_ID FROM Paragraphs GROUP BY Document\_ID HAVING count(*) BETWEEN 1 AND 2} \\
    \bottomrule
    \end{tabularx}%
  \caption{Selected Examples.}
  \label{tab:errorana}%
  \vspace{-3mm}
\end{table*}%

We further select examples from the \textsc{Spider} development set, presented in Table~\ref{tab:errorana}, to show the improved prediction of our model.
The \textit{baseline} system refers to RAT-SQL+BART Encoder model and \textit{our model} refers to the RAT-SQL+\modelnamelm~Encoder.
Overall, our model achieve better column selection performance, either explicit matching between the schema and utterance~(e.g. in \textit{Example 1}, \textit{how much does it weigh} should match \texttt{weight} instead of \texttt{pet\_age}), or implicit matching~(e.g. in \textit{Example 4}, \textit{arriving in ASY Airport} should match \texttt{DestAirport} instead of \texttt{Airline}).
Furthermore, our model can handle complex question better~(e.g. in \textit{Example 5}, our model can generate \texttt{HAVING avg(LifeExpectancy) < 72} condition).

\end{document}